\documentclass[runningheads]{llncs}
\usepackage{amsmath}
\usepackage{amssymb}
\usepackage{amsfonts}
\usepackage{tabularx}
\usepackage{booktabs}
\usepackage{multirow}
\usepackage{pifont}

\usepackage[T1]{fontenc}
\usepackage{graphicx}
\begin{document}
\title{Visual Grounding of Whole Radiology Reports for 3D CT Images}
\titlerunning{Visual Grounding of Radiology Reports for CT Images}
%
\author{Akimichi Ichinose\inst{1} \and
Taro Hatsutani\inst{1} \and
Keigo Nakamura\inst{1} \and
Yoshiro Kitamura\inst{1} \and
Satoshi Iizuka\inst{2} \and
Edgar Simo-Serra\inst{3} \and
Shoji Kido\inst{4} \and
Noriyuki Tomiyama\inst{4}
}

\authorrunning{A. Ichinose et al.}

\institute{
Medical Systems Research \& Development Center, FUJIFILM Corporation, Japan \and
Center for Artificial Intelligence Research, University of Tsukuba, Japan \and
Department of Computer Science and Engineering, Waseda University, Japan \and
Graduate School of Medicine, Osaka University, Japan
\email{akimichi.ichinose@fujifilm.com}
}

\maketitle              

\begin{abstract}
Building a large-scale training dataset is an essential problem in the development of medical image recognition systems.
Visual grounding techniques, which automatically associate objects in images with corresponding descriptions, can facilitate labeling of large number of images.
However, visual grounding of radiology reports for CT images remains challenging, because so many kinds of anomalies are detectable via CT imaging, and resulting report descriptions are long and complex.
In this paper, we present the first visual grounding framework designed for CT image and report pairs covering various body parts and diverse anomaly types. 
Our framework combines two components of 1) anatomical segmentation of images, and 2) report structuring. 
The anatomical segmentation provides multiple organ masks of given CT images, and helps the grounding model recognize detailed anatomies. 
The report structuring helps to accurately extract information regarding the presence, location, and type of each anomaly described in corresponding reports. 
Given the two additional image/report features, the grounding model can achieve better localization.
In the verification process, we constructed a large-scale dataset with region-description correspondence annotations for 10,410 studies of 7,321 unique patients.
We evaluated our framework using grounding accuracy, the percentage of correctly localized anomalies, as a metric and demonstrated that the combination of the anatomical segmentation and the report structuring improves the performance with a large margin over the baseline model (66.0\% vs 77.8\%). 
Comparison with the prior techniques also showed higher performance of our method.

\keywords{Deep Learning \and Vision Language \and Visual Grounding \and Computed Tomography.}
\end{abstract}

\section{Introduction}

\begin{figure}
\centering
\includegraphics[width=0.95\textwidth]{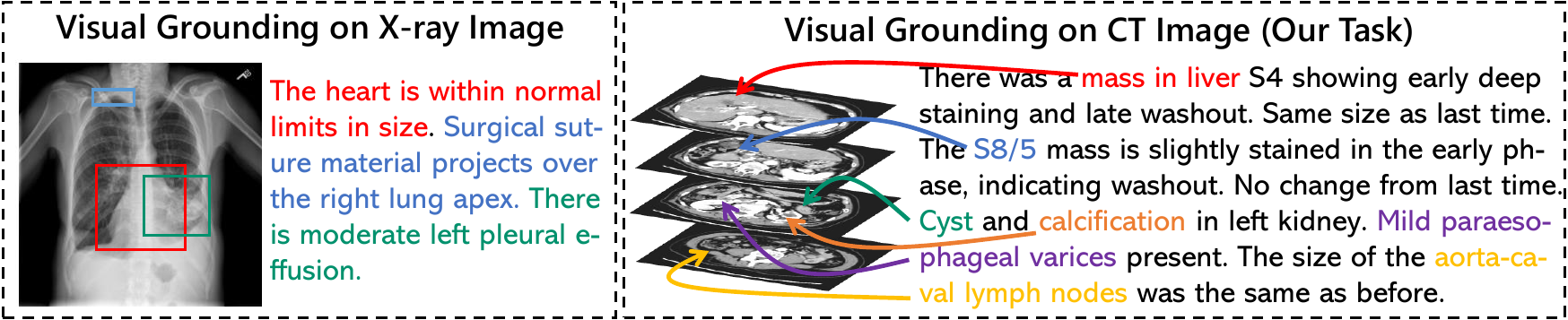}
\caption{Comparison of the visual grounding task on X-ray image and on CT image.} \label{Taskfig}
\end{figure}

In recent years, a number of medical image recognition systems have been developed~\cite{Ebrahimian2022} to alleviate the increasing burden on radiologists~\cite{Dall2020,Rimmer2017,Nishie2015}.
In the development of such systems, the task of manually labeling images is a significant bottleneck.
Auto-labeling, the process of automatically assigning labels to images using machine learning algorithms, has emerged as a promising solution to this problem.
In cases where there are plenty of image and caption pairs, one potential approach to auto-labeling is visual grounding~\cite{Karpathy2014}, which utilizes natural language descriptions to identify and localize objects in images.

With the recent advances in cross-modal technology based on deep learning, many frameworks for visual grounding has been proposed~\cite{Karpathy2015,Hu2016}.
Within the medical domain, several large scale datasets with radiology reports are available (e.g. OpenI~\cite{Demner2016}, MIMIC-CXR~\cite{Johnson2019}), and these produced researches on medical image visual grounding~\cite{Wang2018,Bhalodia2021}.
However, to the best of our knowledge, prior studies have focused on 2D X-ray images~\cite{You2021} or videos~\cite{Li2022}, and there has been no research applying visual grounding to 3D computed tomography (CT) images so far.
Visual grounding on CT images has the following difficulties:
1) \textbf{Large number of anomaly types to detect}: 
Existing researches on visual grounding using X-ray images handled only chest X-ray images. 
The number of anomaly types to detect is at most dozen or so (e.g. 13 findings~\cite{Irvin2019}).
In contrast, our research handles CT images including various parts of the human body. 
Consequently, the number of anomaly types to be detected is larger than one hundred.
2) \textbf{Long and complex sentences}: 
Radiology reports on X-ray images are often simple, noting only the presence or absence of anomalies.
On the other hand, in CT examinations, the qualitative diagnosis of each anomaly is often performed. 
In cases, multiple anomalies are simultaneously described in a sentence.
Therefore, the description tend to be long and complicated with multiple sentences (Fig.~\ref{Taskfig}).
Visual grounding for CT images requires the extraction of information about the location and type of each anomaly from these complex sentences.

In this work, we propose a novel visual grounding framework for 3D CT images and radiology reports.
The main idea is to separate the task into three parts: 1) anatomical segmentation on images, 2) report structuring, and 3) localization of described anomalies.
In the anatomical segmentation, multiple organs and tissues are extracted using the deep learning based segmentation model and provided as landmarks.
The report structuring model, which is based on BERT~\cite{Devlin2019}, is also introduced to extract information of each anomaly from a complex report.
Both of these features are fed into the grounding model (3) to extrapolate medical domain knowledge, thereby enabling accurate visual grounding.

Our contributions are as follows:
\begin{itemize}
\item We show the first visual grounding results for 3D CT images that covers various body parts and anomalies.
\item We introduce a novel grounding architecture that can leverage report structuring results of presence/type/location of described anomalies.
\item We validate the efficacy of the proposed framework using a large-scale dataset with region-description correspondence annotations.
\end{itemize}

\section{Related Work}
\paragraph{\textbf{Visual Grounding}}
Visual grounding task involves learning the correspondences between descriptions in the text and image regions from a given training set of region-description pairs~\cite{Karpathy2014}. 
There are mainly two approaches: one-stage approach and two-stage approach.
Most studies follow a two-stage approach~\cite{Lee2018,Lu2019}.
However, this approach usually employs a pre-trained object detector, and it leads to restrict the capability of categories and attributes in grounding. 
Accordingly, recent studies is shifting to employ the one-stage approach, in which visual grounding is performed by end-to-end training~\cite{Yang2019,Deng2021,Kamath2021}.

\paragraph{\textbf{Vision-Language Tasks on Medical Image}}
The existence of public datasets with paired images and reports~\cite{Demner2016,Johnson2019,Yan2018} has accelerated research on cross-modal tasks in the medical field~\cite{Wang2018,Li2020}. 
Inspired by the success of visual grounding, several studies of visual grounding for medical images and radiology reports have also been reported~\cite{You2021,Bhalodia2021,Seibold2022}.
These studies utilized a large scale dataset and an attention-based language interpretation model such as BERT~\cite{Devlin2019} to ground the descriptions in the report.
However, these studies have focused on X-ray images, and to the best of our knowledge, there have been no studies on CT images, which cover the entire body and have a complex report.

\section{Methods}
We first formulate the problem. 
Next, we explain three key components of anatomical segmentation, report structuring, and anomaly localization in our framework.
In our framework, multiple organ labels obtained as the output of anatomical segmentation encourage the grounding model to learn detailed anatomy, and report structuring allows the grounding model to accurately extract the features of the target anomaly from complex sentences.

\subsection{Problem Formulation}
\begin{figure}
\centering
\includegraphics[width=0.9\textwidth]{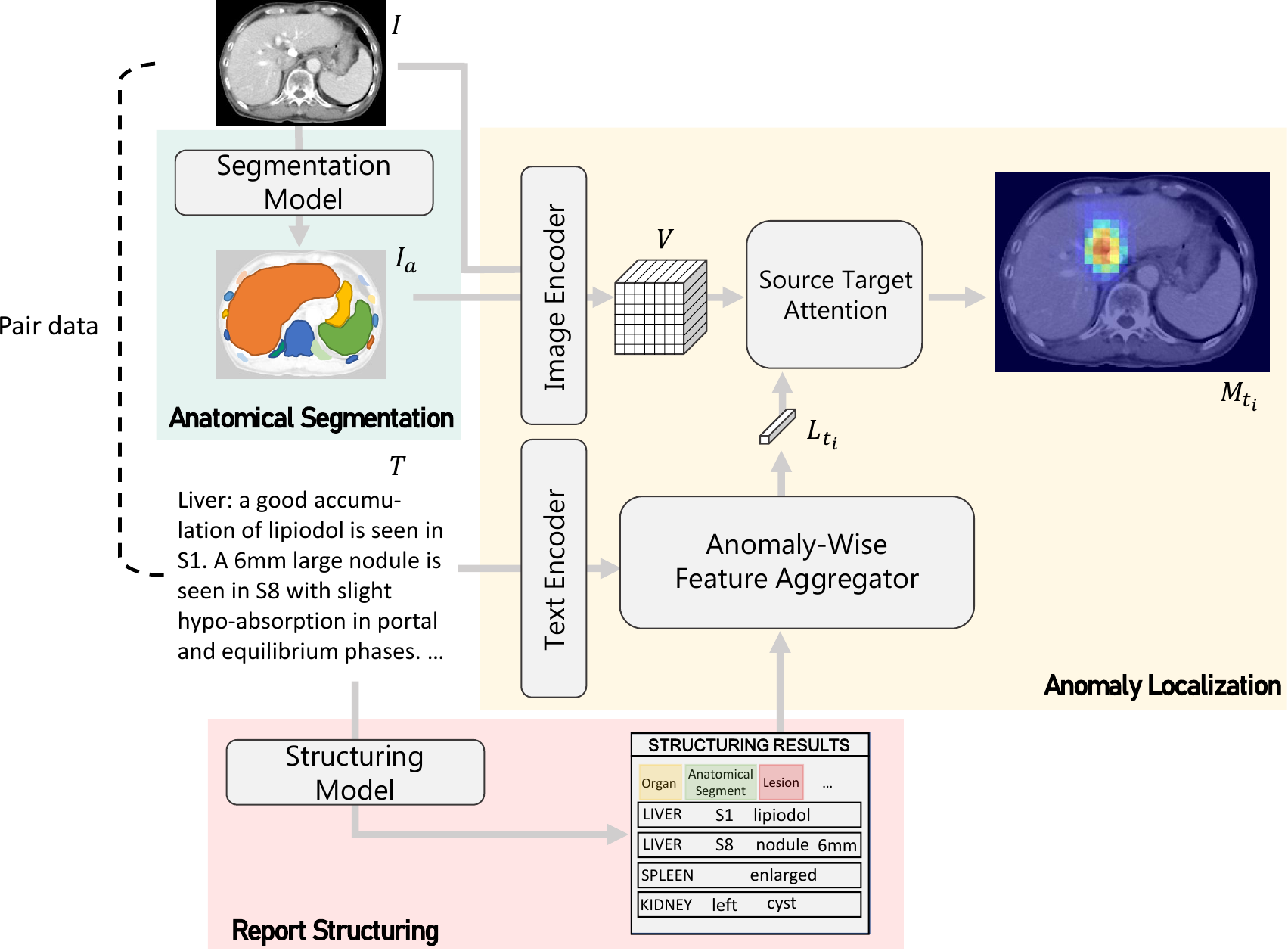}
\caption{The proposed framework for 3D-CT visual grounding.} \label{OverallArchitecture}
\end{figure}

Our research assumes that a dataset of image-report pairs with region-description correspondence annotations is provided for training.
We show the overall framework in Fig.~\ref{OverallArchitecture}.
We denote an image and a paired report as \textit{I} and \textit{T} respectively.
Let $I_a$ be a label image in which multiple organs are extracted from $I$.
Each report $T$ contains descriptions of multiple (image) anomalies.
We denote each anomalies as $t_i \in \{ t_1, t_2, ..., t_{N} \}$. 
Given an image \textit{I} and corresponding organ label images $I_a$ encoded as $V\in\mathbb{R}^{d_z \times d_y \times d_x \times d}$ and a description about an anomaly $t_i$ encoded as $L_{t_i}\in\mathbb{R}^{d}$, the goal of our framework is to generate a segmentation map $M_{t_i}$ that represents the location of the anomaly $t_i$. 

\subsection{Anatomical Segmentation}
The task of the anatomical segmentation is to extract relevant anatomies that can be clues for visual grounding.
We use the commercial version of the 3D image analysis software (Synapse 3D V6.8, FUJIFILM corporation, Japan) to extract 32 organs and tissues (See Appendix Table.~A1). 
In this software, anatomies are extracted using U-Net based architectures~\cite{Keshwani2018,Masuzawa2020}.
The extracted anatomical label images are $I_a$.

\subsection{Report Structuring}
The tasks of the report structuring are as follows: 1) anatomical prediction, 2) phrase recognition, and 3) relationship estimation between phrases (See Appendix Fig.~A1). 
The anatomical prediction is a sentence-wise prediction to determine which organ or body part is mentioned in each sentence. 
The organs and body parts to be recognized are shown in Appendix Table.~A2.
The sentences belonging to the same class are concatenated, then the phrase recognition and the relationship estimation are performed for each class.

The phrase recognition module extracts phrases and classifies each of them into 9 classes (See Appendix Table.~A2). 
Subsequently, the relationship estimation module determines whether there is a relationship between anomaly phrases (e.g. 'nodule', 'fracture') and other phrases (e.g. '6mm', 'Liver S6'), resulting in the grouping of phrases related to the same anomaly.
If multiple anatomical phrases are grouped in the same group, they are split into separate groups on a rule basis (e.g. [‘right S1’, ‘left S6’, ‘nodule’] -> [‘right S1’, ‘nodule’], [‘left S6’, ‘nodule’]).
More details of implementation and training methods are reported in Nakano et al.~\cite{nakano2022} and Tagawa et al~\cite{tagawa2022}.

\subsection{Anomaly Localization}
The task of the anomaly localization is to output a localization map of the anomaly mentioned in the input report $T$. 
The CT image $I$ and the organ label image $I_a$ are concatenated along the channel dimension and encoded by a convolutional backbone to generate a visual embedding $V$.
The sentences in the report $T$ are encoded by BERT~\cite{Devlin2019} to generate embeddings for each character.
Let $r = \{r_{1}, r_{2}, ..., r_{N_C}\}$ be the set of character embeddings where $N_C$ is the number of characters.
Our framework next adopt the Anomaly-Wise Feature Aggregator (AFA).
For each anomaly $t_i$, AFA generates a representative embedding $L_{t_i}$ by aggregating the embeddings of related phrases based on report structuring results.
The final grounding result $M_{t_i}$ is obtained by the following Source-Target Attention.

\begin{equation}
M_{t_i} = \text{sigmoid}(L_{t_i}W_Q(VW_K)^T)
\end{equation}
where $W_Q$, $W_K$ $\in$ $\mathbb{R}^{d \times d_n}$ are trainable variables.

The overall architecture of this module is illustrated in Appendix Fig.~A2.

\subsubsection{Anomaly-Wise Feature Aggregator}
The results of the report structuring $m_{t_i} \in \mathbb{R}^{N_C}$ are defined as follows:
\begin{equation}
m_{t_{ij}} =
  \begin{cases}
    c_j& \text{if a $j$-th character is related to an anomaly $t_i$,} \\
    0 & \text{else.}
  \end{cases}
\end{equation}
\begin{equation}
m_{t_i} = \{m_{t_{i1}}, m_{t_{i2}}, ... m_{t_{iN_C}}\}
\end{equation}
where $c_j$ is the class index labeled by the phrase recognition module (Let $C$ be the number of classes).
In this module, aggregate character-wise embeddings based on the following formula.
\begin{equation}
e_k = \{r_j | m_{t_{ij}} = k\}
\end{equation}
\begin{equation}
L_{t_i} = \text{LSTM}([v_{organ};p_1;e_1;p_2;e_2;...,p_C;e_C])
\end{equation}
where $v_{organ}$ and $p_k$ are trainable embeddings for each organ and each class label respectively.
$[\cdot ; \cdot]$ stands for concatenation operation.
In this way, embeddings of characters related to the anomaly $t_i$ are aggregated and concatenated.
Subsequently, representative embeddings of the anomaly are generated by an LSTM layer.
In the task of visual grounding focused on 3D CT images, the size of the dataset that can be created is relatively small.
Considering this limitation, we use an LSTM layer with strong inductive bias to achieve high generalization performance.

\section{Dataset and Implementation Details}

\subsection{Clinical Data}
We retrospectively collected 10,410 CT studies (11,163 volumes/7,321 unique patients) and 671,691 radiology reports from one university hospital in Japan. 
We assigned a bounding box to each anomaly described in the reports as shown in Appendix Fig.~A3.
The total category number is about 130 in combination of anatomical regions and anomaly types (The details are in Fig. ~\ref{WRfig}) 
For each anomaly, a correspondence annotation was made with anomaly phrases in the report.
The total number of annotated regions is 17,536 (head: 713 regions, neck: 285 regions, chest: 8,598 regions, and abdomen: 7,940 regions).
We divide the data into 9,163/1,000/1,000 volumes as a training/validation/test split.

\subsection{Implementation Details}
We use a VGG-like network as Image Encoder, with 15 3D-convolutional layers and 3 max pooling layers.
For training, the voxel spacings in all three dimensions are normalized to 1.0 mm. 
CT values are linearly normalized to obtain a value of [0--1].
The anatomy label image, in which only one label is assigned to each voxel, is also normalized to the value [0--1], and the CT image and the label image are concatenated along the channel dimension.
As our Text Encoder, we use a BERT with 12 transformer encoder layers, each with hidden dimension of 768 and 12 heads in the multi-head attention.
At first, we pre-train the BERT using 6.7M sentences extracted from the reports in a Masked Language Model task.
Then we train the whole architecture jointly using dice loss~\cite{Milletari2016} with the first 8 transformer encoder layers of the BERT frozen.
Further information about implementation are shown in Appendix Table.~A3.

\section{Experiments}
We did two kinds of experiments for comparison and ablation studies.
The comparison study was made against TransVG~\cite{Deng2021} and MDETR~\cite{Kamath2021} that are one-stage visual grounding approaches and established state-of-the-art performances on photos and captions.
To adapt TransVG and MDETR for the 3D modality, the backbone was changed to a VGG-like network with 3D convolution layers, the same as the proposed method.
We refer one of the proposed method without anatomical segmentation and report structuring as the baseline model.

\subsection{Evaluation Metrics}
We report segmentation performance using Dice score, mean intersection over union (mIoU), and the grounding accuracy.
The output masks are thresholded to compute mIoU and grounding accuracy score.
The mIoU is defined as an average IoU over the thresholds [0.1, 0.2, 0.3, 0.4, 0.5]. 
The grounding accuracy is defined as the percentage of anomalies for which the IoU exceeds 0.1 under the threshold 0.1.

\subsection{Results}

The experimental results of the two studies are shown in Table.~\ref{Restab}.
Both of MDETR and TransVG failed to achieve stable grounding in this task.
A main difference between these models and our baseline model is using a source-target attention layer instead of the transformer.
It is known that a transformer-based algorithm with many parameters and no strong inductive bias is difficult to generalize with such a relatively limited number of training data.
For this reason, the baseline model achieved a much higher accuracy than the comparison methods.

\newcolumntype{D}{>{\centering\arraybackslash}p{20mm}}
\newcolumntype{E}{>{\centering\arraybackslash}p{15mm}}
\newcommand{\cmark}{\ding{51}}%
\newcommand{\xmark}{\ding{55}}%
\begin{table}
\centering
\caption{Results of the comparison/ablation studies. '-' represents 'not converged'.}\label{Restab}
\begin{tabular}{lDDEEE}
\toprule
\multirow{2}{*}{\textbf{Method}} & \textbf{Anatomical} & \textbf{Report} & \textbf{Dice} & \textbf{mIoU} & \textbf{Accuracy} \\
 & \textbf{Seg.} & \textbf{Struct.} & $[\%]$ & $[\%]$ & $[\%]$ \\
\midrule
MDETR~\cite{Kamath2021} & - & - & N/A & - & - \\
TransVG~\cite{Deng2021} & - & - & N/A & 8.5 & 21.8 \\
\midrule
Baseline & \xmark & \xmark & 27.4 & 15.6 & 66.0 \\  
\multirow{3}{*}{Proposed} & \cmark & \xmark & 28.1 & 16.6 & 67.9 \\  
 & \xmark & \cmark & 33.0 & 20.3 & 75.9 \\  
 & \cmark & \cmark & \textbf{34.5} & \textbf{21.5} & \textbf{77.8} \\  
\bottomrule
\end{tabular}
\end{table}

\begin{figure}
\centering
\includegraphics[width=0.990\textwidth]{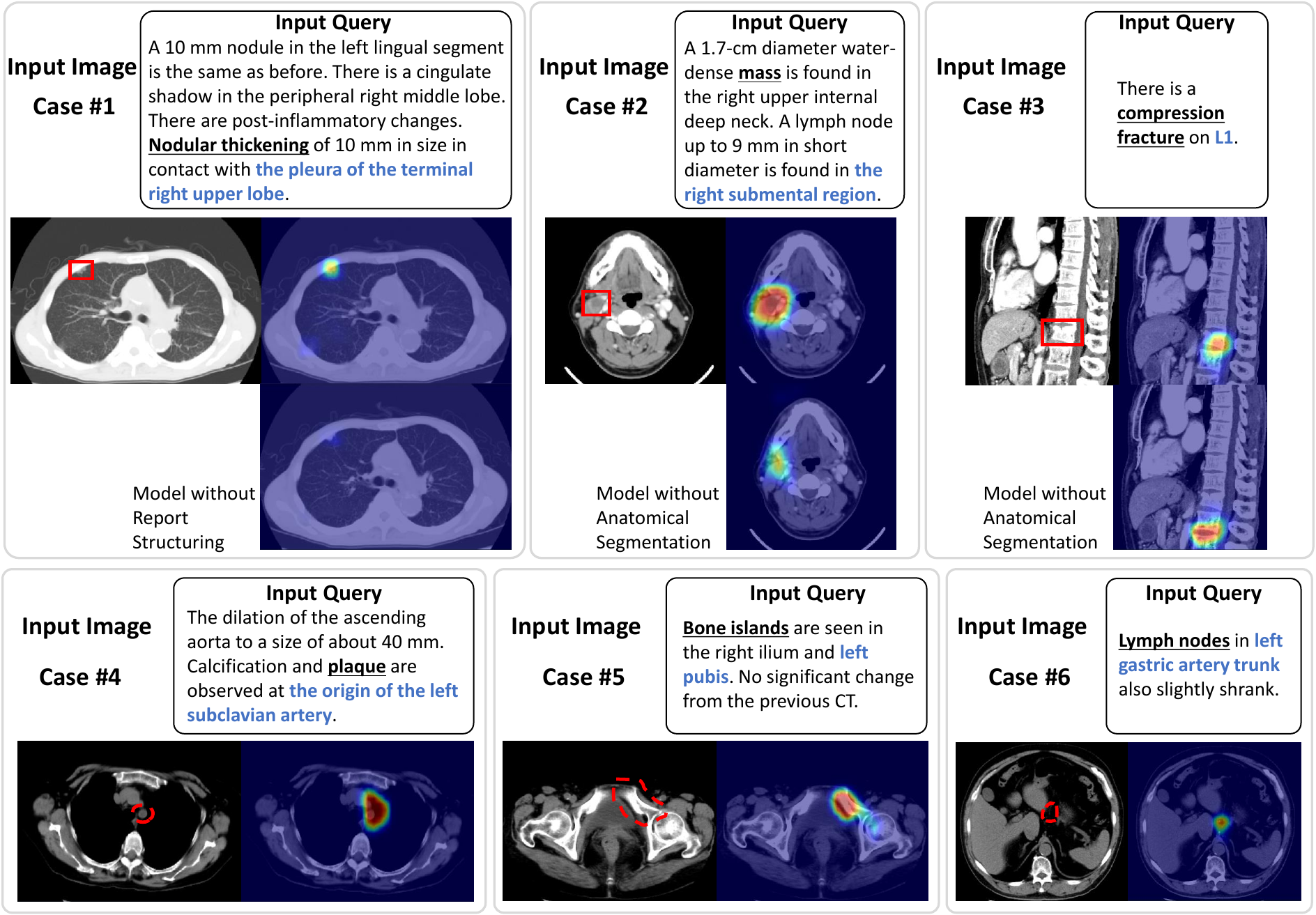}
\caption{The grounding results for several input queries. Underlines in the input query indicate the target anomaly phrase to be grounded. The phrases highlighted in bold blue indicate the anatomical locations of the target anomaly. The red rectangles indicate the ground truth regions. Case \#4-\#6 are the grounding results when an unrelated input query is inputted. The region surrounded by the red dashed line indicates the anatomical location corresponding to the input query.} \label{GRfig}
\end{figure}

The ablation study showed that the anatomical segmentation and the report structuring can improve the performance.
In Fig.~\ref{GRfig} (upper row), we demonstrate several cases that facilitate an intuitive understanding of each effect.
Longer reports often mention more than one anomaly, making it difficult to recognize the grounding target and cause localization errors.
The proposed method can explicitly indicate phrases such as the location and size of the target anomaly, reducing the risk of failure.
Fig.~\ref{GRfig} (lower row) shows examples of grounding results when a query that is not related to the image is inputted. 
In this case, the grounding results were less consistent with the anatomical phrases.
The results suggest that the model performs grounding with an emphasis on anatomical information against the backdrop of abundant anatomical knowledge.

\begin{figure}
\centering
\includegraphics[width=\textwidth]{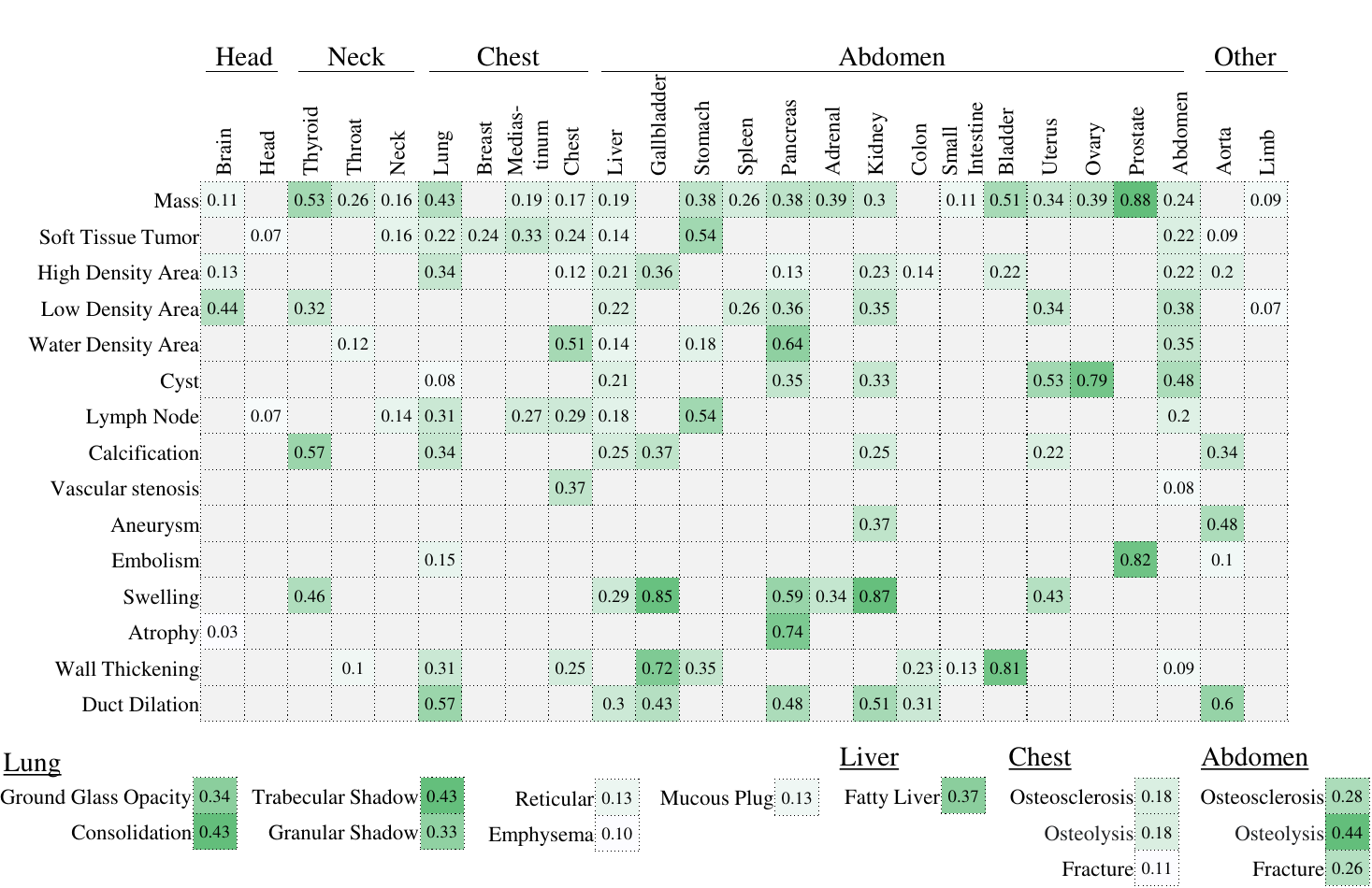}
\caption{Grounding performance for representative anomalies. The value in each cell is the average dice score of the proposed method.} \label{WRfig}
\end{figure}

The grounding performance for each combination of organ and anomaly type is shown in Fig.~\ref{WRfig}.
The performance is relatively high for organ shape abnormalities (e.g. swelling, duct dilation) and high-frequency anomalies in small organs (e.g. thyroid/prostate mass).
For these anomaly types, our model is considered to be available for automatic training data generation.
On the other hand, the performance tends to be low for rare anomalies (e.g. mass in small intestine) and anomalies in large body part (e.g. limb).
Improving grounding performance for these targets will be an important future work.

\section{Conclusion}
In this paper, we proposed the first visual grounding framework for 3D CT images and reports. 
To deal with various type of anomalies throughout the body and complex reports, we introduced a new approach using anatomical recognition results and report structuring results.
The experiments showed the effectiveness of our approach and achieved higher performance compared to prior techniques.
However, in clinical practice, radiologists write reports from comparing multiple images such as time-series images, or multi-phase scans.
Realizing such sophisticated diagnose process by a visual grounding model will be a future research.

%
%
\bibliographystyle{splncs04}
\bibliography{paper2226}

\newpage
\setcounter{table}{0}
\renewcommand{\thetable}{A\arabic{table}}

\setcounter{figure}{0}
\renewcommand{\thefigure}{A\arabic{figure}}

\begin{figure}
\centering
\includegraphics[width=0.7\textwidth]{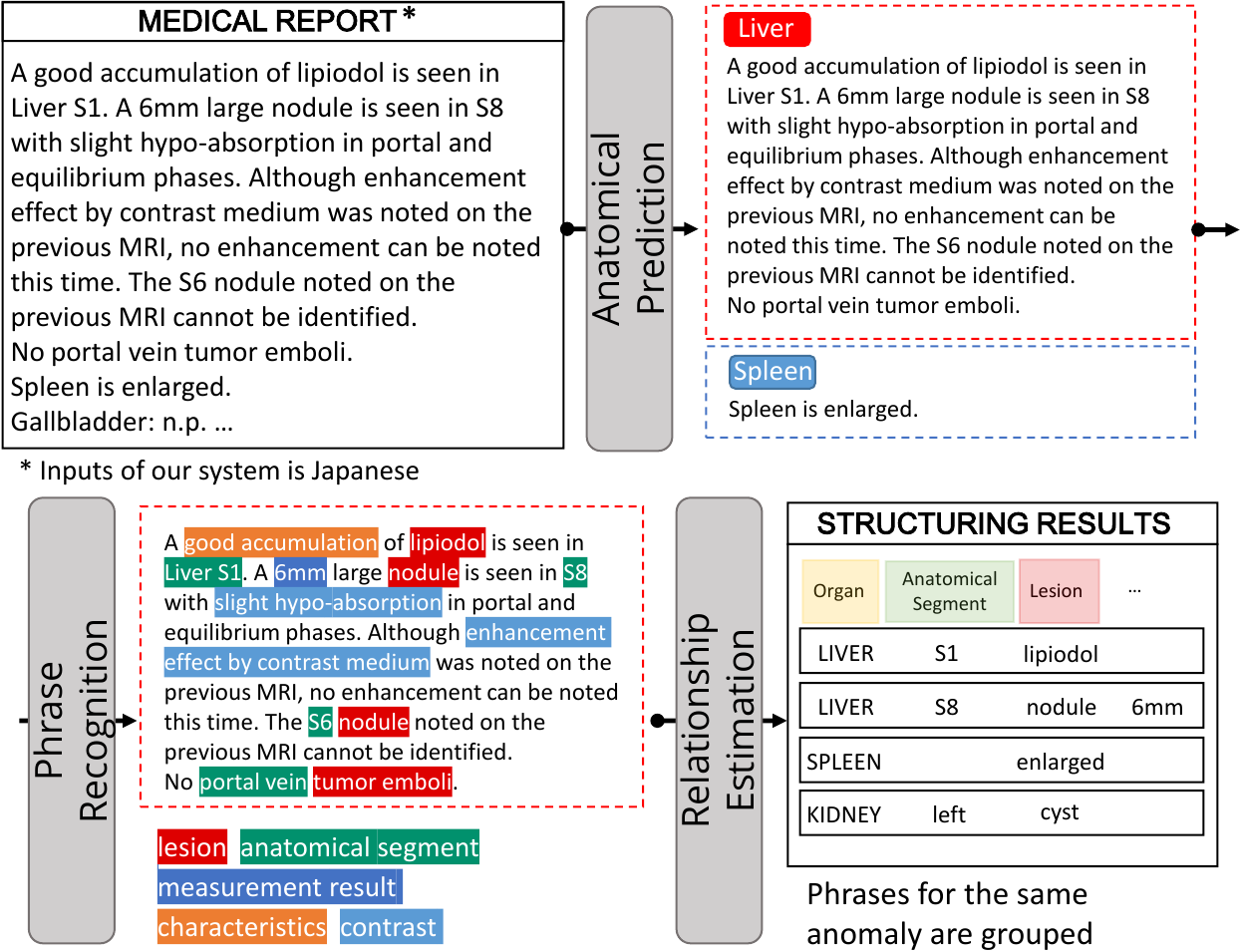}
\caption{Flowchart of report structuring.} \label{A-RSfig}
\end{figure}

\samepage
\newcolumntype{D}{>{\arraybackslash}p{35mm}}
\begin{table}
\centering
\caption{Output classes of anatomical segmentation}\label{A-OrganSeg}
\begin{tabular}{llll}
\toprule
\quad Brain  & \quad Lung & \quad Liver & \quad Aorta \\
\quad- Cerebraspinal Fluid & \quad- Left Upper Lobe & \quad Gallbladder & \quad Bone \\
\quad- Left Lateral Ventricle &  \quad- Left Lower Lobe & \quad Stomach &  \quad- Cervical Vertebra\\
\quad- Right Lateral Ventricle & \quad- Right Upper Lobe & \quad Duodenum &  \quad- Thoracic Vertebra \\
\quad- Third Ventricle & \quad- Right Middle Lobe & \quad Spleen &  \quad- Lumber Vertebra  \\
\quad- Fourth Ventricle & \quad- Right Lower Lobe & \quad Pancreas  & \quad- Left Rib \\
\quad- Brainstem & \quad Heart & \quad Kidney & \quad- Right Rib \\
\quad- Left Cerebellum &  & \quad- Left Kidney & \\
\quad- Right Cerebellum &  & \quad- Right Kidney&  \\
\quad- Left Cerebrum & & \quad Prostate & \\
\quad- Right Cerebrum  & & \quad Bladder & \\
\bottomrule
\end{tabular}
\end{table}

\samepage
\newcolumntype{E}{>{\arraybackslash}p{25mm}}
\newcolumntype{F}{>{\arraybackslash}p{30mm}}
\begin{table}
\centering
\caption{Output classes of report structuring}\label{A-ReportStr}
\begin{tabular}{EEFD}
\toprule
\multicolumn{3}{c}{Anatomical Prediction} & Phrase Classification\\
\midrule
Head & Breast & Small Intestine & Anatomical Segment \\
Brain & Pleural Cavity & Colon & Lesion \\
Ear & Chest & Prostate & Shape Abnormality \\
Nose & Liver & Uterus & Diagnosis \\
Neck & Stomach & Bladder & Characteristics \\
Oropharynx & Gallbladder & Ovary & Contrast Information \\
Thyroid & Adrenal & Abdomen & Quantity \\
Lung & Spleen & Abdominal Cavity & Measurement Result \\
Heart & Pancreas & Limb & Temporal Change \\
Mediastinum & Kidney & Aorta &  \\
\bottomrule
\end{tabular}
\end{table}

\begin{figure}
\centering
\includegraphics[width=0.66\textwidth]{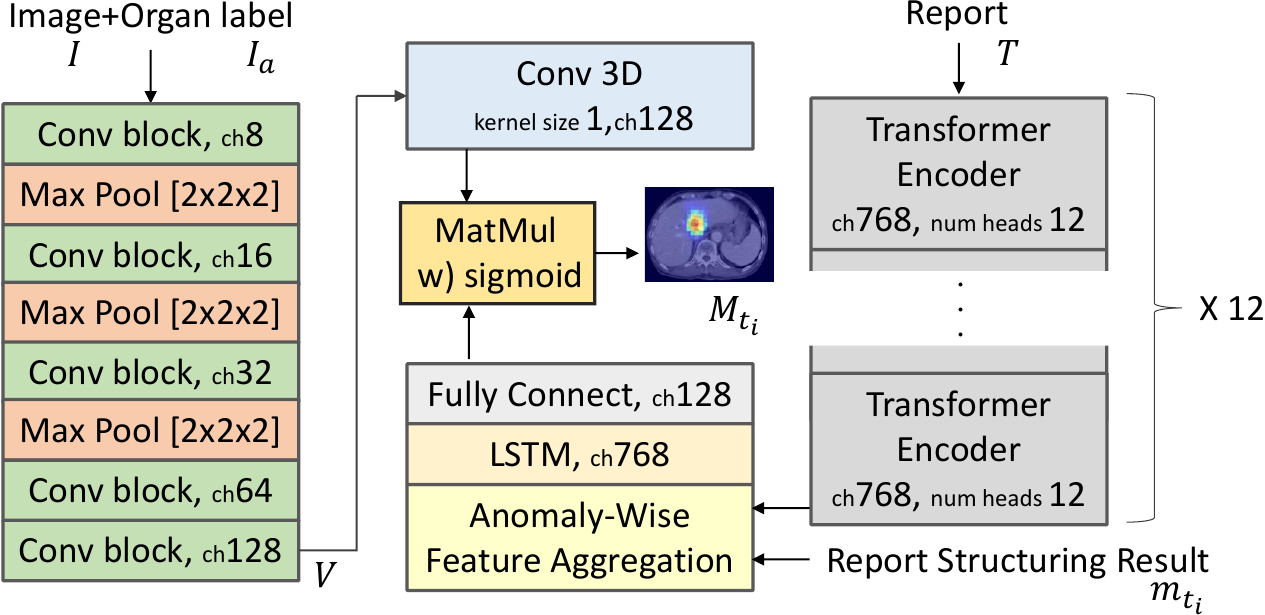}
\caption{Anomaly localization architecture: The Conv block denotes three sets of $Conv[3 \times 3 \times 3] \rightarrow BatchRenorm \rightarrow ReLU$ operations. } \label{A-ALModelfig}
\end{figure}

\begin{figure}
\centering
\includegraphics[width=0.8\textwidth]{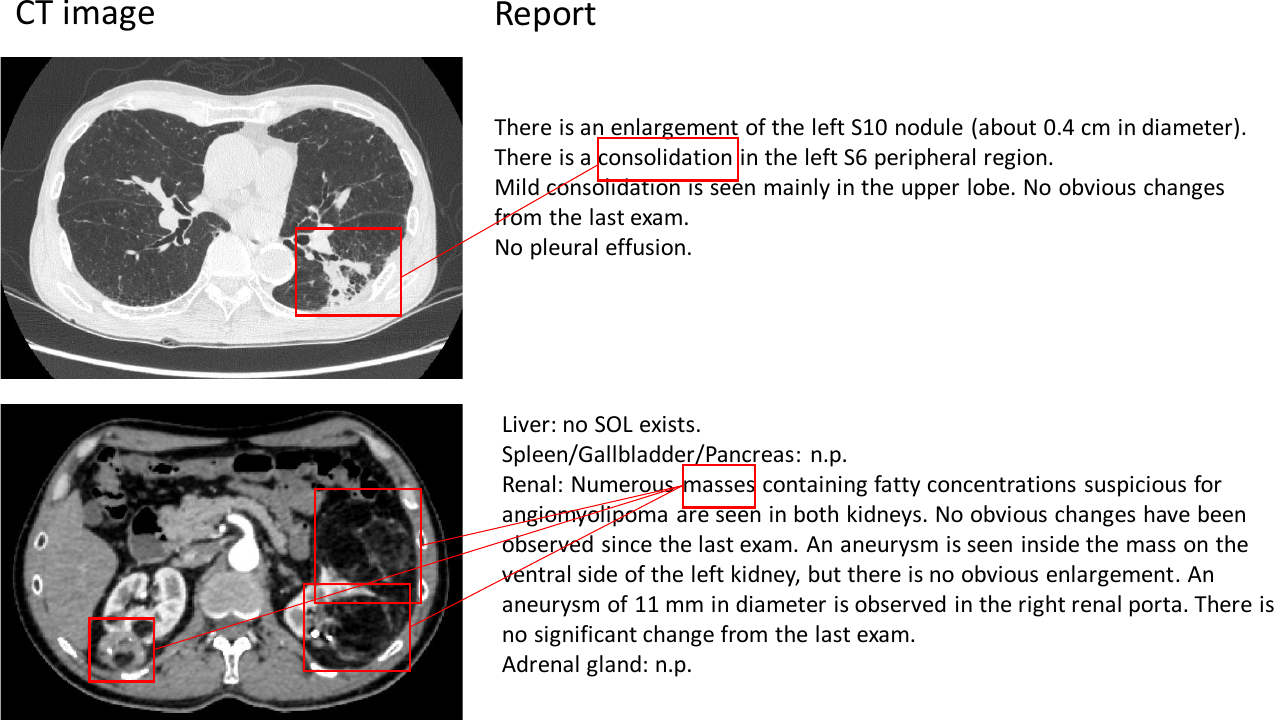}
\caption{Examples of the ground truth data.} \label{A-GT}
\end{figure}
\newcolumntype{G}{>{\arraybackslash}p{70mm}}
\begin{table}
\centering
\caption{Implementation Details}\label{A-ID-Table}
\begin{tabular}{l|G}
\toprule
 & \multicolumn{1}{c}{\textbf{Value}} \\
\midrule
Optimizer & Adam \\
Initial Learning Rate & $1e^{-5}$ \\
Learning Rate Schedule & Linearly increased to $1e^{-4}$ within the first 5,000 steps, and then multiplied by 0.1 every 30,000 steps \\
Batch Size & 10 \\
Normalization Method &  Batch Renormalization\\
Data Augmentation for Image & Random Crop, Random Rotation, Random Scaling, Sharpness change, Smoothing, and Gaussian noise addition\\
Data Augmentation for Text & Random deletion, Random insertion, and Random crop\\
Machine Learning Library & Tensorflow 2.3 \\
GPU & NVIDIA Tesla V100 $\times$ 2\\
\bottomrule
\end{tabular}
\end{table}

\end{document}